\documentclass[11pt]{article}

\usepackage[a4paper,margin=1in]{geometry}
\usepackage{times}
\usepackage{graphicx}
\usepackage{amsmath}
\usepackage{booktabs}
\usepackage{hyperref}

\title{\textbf{Modeling Authorial Style in Urdu Novels Using Character Interaction Graphs and Graph Neural Networks}}

\author{
Hamza Naveed\\
Hassan Mujtaba\\
Hanzlah Munir\\
Department of Computer Science\\
Information Technology University, Lahore\\
\texttt{bscs23174@itu.edu.pk}
}

\date{}

\begin{document}
\maketitle

\begin{abstract}
Authorship analysis has traditionally relied on lexical and stylistic cues derived from textual content. However, narrative structure---particularly patterns of character interactions---also encodes distinctive authorial traits that remain largely unexplored, especially for low-resource languages such as Urdu. In this work, we propose a graph-based framework for modeling Urdu novels as character interaction networks and investigate whether authorial style can be inferred from these structures alone. Each novel is converted into a graph where nodes represent characters and edges represent co-occurrence within narrative proximity. We conduct a systematic comparison of multiple graph representations, including global structural features, node-semantic summaries, unsupervised graph embeddings, and supervised graph neural networks. Experiments on 52 Urdu novels from seven authors demonstrate that learned graph representations significantly outperform hand-crafted and unsupervised baselines, achieving up to 0.857 accuracy under a strict author-aware evaluation protocol.
\end{abstract}

\section{Introduction}

Authorship attribution has traditionally focused on lexical, syntactic, and stylistic features extracted directly from text \cite{stamatatos2009}. While such approaches have been successful, they largely ignore the underlying narrative structure through which stories are organized.

In character-driven literature, authors implicitly construct social systems where characters interact, form relationships, and influence the progression of the narrative. Prior work has shown that literary texts can be modeled as social networks, enabling the study of narrative structure through graph-based analysis \cite{moretti2011,elson2010}.

Urdu literature, despite its richness, remains underexplored in computational literary analysis due to limited digital resources and language-specific challenges. This motivates the exploration of structure-based approaches that are less dependent on linguistic features.

In this work, we investigate whether authorial style in Urdu novels can be captured using character interaction graphs alone, without relying on textual content. We systematically evaluate multiple graph representations and demonstrate the effectiveness of learned graph embeddings under careful, leakage-safe evaluation.
\newpage
\section{Related Work}

\subsection{Authorship Attribution}
Authorship attribution has been extensively studied using stylometric and statistical features derived from text \cite{stamatatos2009}. These methods are effective but sensitive to vocabulary, genre, and language.

\subsection{Social Network Analysis in Literature}
The idea of modeling literary narratives as networks of characters has been explored in both literary theory and computational studies. Moretti \cite{moretti2011} and Elson et al. \cite{elson2010} demonstrated that character interaction networks capture meaningful narrative structure.

\subsection{Graph Representation Learning}
Graph representation learning methods provide ways to encode graph structure into fixed-length vectors. Unsupervised methods such as Graph2Vec \cite{narayanan2017graph2vec} capture global structure, while Graph Neural Networks enable supervised learning over both structure and node attributes \cite{kipf2017semi,velickovic2018graph}.

\section{Data Collection and Graph Construction}

Our dataset consists of 52 Urdu novels written by seven authors. Many novels were available only as scanned documents and were converted to text using OCR. Texts were segmented into fixed-size portions to ensure consistent processing.

Characters were identified using LLM-assisted extraction followed by normalization of aliases and honorifics. Each novel was represented as an undirected weighted graph where nodes correspond to characters and edges represent co-occurrence within narrative proximity.

Each character node was annotated with basic semantic attributes such as gender and narrative role.

\section{Methodology}

This section describes how Urdu novels are transformed into graph representations and how different graph-level features and models are used for authorship analysis. Our methodology is designed to progressively increase representational expressiveness while maintaining interpretability and evaluation fairness.

\subsection{Graph Construction}

Each novel is modeled as a character interaction graph. Nodes represent unique characters appearing in the narrative, and edges represent narrative proximity based on character co-occurrence. Two interaction definitions are explored: (1) co-occurrence within the same page and (2) co-occurrence within a fixed window of five pages. Edge weights correspond to the frequency of such co-occurrences across the novel.

This formulation provides an objective and reproducible approximation of character interactions without requiring subjective interpretation of dialogue or action.
\newpage
\subsection{Node Attribute Encoding}

To incorporate semantic information, each character node is annotated with a small set of interpretable attributes:
\begin{itemize}
    \item \textbf{Gender}: male, female, or unknown
    \item \textbf{Narrative role}: protagonist, antagonist, support, narrator, or minor
\end{itemize}

These attributes are encoded using one-hot vectors and serve as node features for graph neural network models. The limited feature set is intentionally chosen to avoid overfitting and ensure interpretability.

\subsection{Graph-Level Representations}

We explore multiple strategies to obtain fixed-length representations of novels from their graph structure.

\paragraph{Structural Features}
Global structural properties such as graph size, density, degree statistics, and clustering coefficients are extracted to form hand-crafted graph-level feature vectors.

\paragraph{Semantic Attribute Summaries}
For each novel, we compute the proportion of nodes belonging to each gender and narrative role category. These proportions form a semantic summary of character composition at the graph level.

\paragraph{Unsupervised Graph Embeddings}
Graph2Vec is used to learn task-agnostic graph embeddings that capture global structural patterns. These embeddings are evaluated using downstream classification to assess their suitability for authorship analysis.

\subsection{Graph Attention Networks}

To jointly model interaction structure and node attributes, we employ Graph Attention Networks (GATs). GATs compute node representations by attending to neighboring nodes, allowing the model to learn which interactions are most informative.

Given a novel graph, node embeddings produced by GAT layers are aggregated using global mean pooling to obtain a fixed-dimensional graph-level embedding. This embedding serves as a learned fingerprint of the novel's narrative structure.

\subsection{Evaluation Protocol}

Due to the limited dataset size, careful evaluation design is critical. We adopt an author-aware evaluation strategy in which exactly one novel per author is held out as a test instance. This ensures that models are evaluated on unseen books from each author and prevents leakage caused by overlapping author samples in training and testing.

For classical models, cross-validation is used where appropriate. For GAT-based models, the author-aware split is fixed throughout training and evaluation.

\subsection{Embedding-Space Augmentation}

To address data scarcity, we introduce an autoencoder-based augmentation strategy applied to learned graph embeddings. An autoencoder is trained solely on training embeddings to learn a compact latent representation of the embedding space.

Synthetic embeddings are generated by sampling the latent space with Gaussian noise and decoding back to the original embedding space. These synthetic samples are combined with real embeddings to train a downstream classifier. Importantly, test embeddings are never used during autoencoder training or augmentation.

\subsection{Downstream Classification}

Logistic Regression is used as the final classifier for evaluating graph-level embeddings. This choice is intentional: a simple linear model serves as a clean probe of embedding quality. High classification performance with Logistic Regression indicates that author-specific information is well captured and linearly separable in the learned representation space.

\section{Experiments and Results}

This section presents a systematic evaluation of different graph-based representations for authorship analysis. The experiments are designed to progressively assess how much authorial signal is captured by increasingly expressive representations, ranging from simple structural summaries to learned graph embeddings.

\subsection{Experimental Setup}

All experiments are conducted on a dataset of 52 novels written by seven authors. Due to the limited size of the dataset, particular care is taken to design evaluation protocols that prevent data leakage and provide realistic estimates of generalization performance.

For classical machine learning models, we use cross-validation where appropriate. For graph neural network experiments, we adopt an author-aware evaluation strategy in which exactly one novel per author is held out as a test instance. This ensures that models are evaluated on unseen books from each author rather than benefiting from multiple samples of the same author in both training and testing.

Classification performance is reported using accuracy, with additional analysis based on confusion matrices and class-wise performance where relevant.

\subsection{Baseline: Structural Graph Features}

We first evaluate whether global structural properties of character interaction graphs are sufficient for distinguishing authors. Each novel is represented using standard graph statistics, including the number of nodes and edges, graph density, and degree-based measures.

Logistic Regression and Random Forest classifiers are trained on these features. The results show that classification accuracy remains close to random guessing, indicating that global structural summaries alone do not capture distinctive authorial patterns. Expanding the edge definition to include character co-occurrence within a five-page window leads to only marginal improvements, suggesting that while interaction locality matters, coarse summaries remain insufficient.

\subsection{Semantic Node Attribute Summaries}

Next, we examine whether semantic information at the character level improves performance. For each novel, we compute the proportion of characters belonging to different genders, narrative roles, and social titles, resulting in a semantic profile of the book.

Compared to structural features, these semantic summaries yield noticeably better performance, particularly when using non-linear classifiers such as Random Forests. This suggests that authors differ more strongly in the composition of characters they employ than in overall graph connectivity patterns. However, this representation still compresses interaction structure heavily and fails to capture relational dynamics.
\newpage
\subsection{Unsupervised Graph Embeddings}

We evaluate Graph2Vec as an unsupervised baseline for learning graph-level representations. Graph2Vec embeddings are followed by Logistic Regression classification.

Despite capturing global structural patterns, Graph2Vec performs close to chance level across folds. This indicates that task-agnostic structural embeddings are insufficient for authorship attribution in literary graphs, likely because they ignore node semantics and narrative context.

\subsection{Graph Attention Networks}

To jointly model interaction structure and character attributes, we employ Graph Attention Networks (GATs). Each character node is represented using one-hot encoded gender and narrative role features. GAT layers compute context-aware node embeddings, which are aggregated via global mean pooling to produce fixed-length graph-level representations.

We explore multiple training and evaluation variants. Na\"{i}ve stratified splits often result in high training accuracy but unstable or inflated test performance, reflecting overfitting and optimistic evaluation. To address this, we adopt a strict author-aware split where one novel per author is permanently held out for testing.

Under this controlled setup, GAT-based models consistently outperform hand-crafted and unsupervised baselines, demonstrating the value of learned representations. However, performance remains limited by the small size of the dataset.

\subsection{Embedding-Space Augmentation and Final Classification}

To mitigate data scarcity, we introduce an autoencoder-based augmentation strategy applied to learned GAT embeddings. An autoencoder is trained solely on training embeddings and used to generate synthetic samples by perturbing the latent space.

These augmented embeddings are used to train a Logistic Regression classifier, which serves as a linear probe of embedding quality. This final pipeline achieves the highest performance, correctly identifying six out of seven authors (accuracy 0.857) under leakage-safe evaluation.

The strong performance of a simple linear classifier indicates that the learned graph embeddings are highly informative and linearly separable with respect to author identity.

\subsection{Summary of Findings}

Across all experiments, a clear trend emerges: as representations move from coarse global summaries to learned graph embeddings that integrate structure and semantics, author identification performance improves substantially. These results highlight the importance of representation learning and careful evaluation design for graph-based literary analysis.

\section{Discussion}

Our findings indicate that authorial style is weakly encoded in global graph statistics but becomes increasingly apparent as richer representations are employed. Learned graph embeddings successfully capture interaction patterns and character semantics that distinguish authors.

\section{Conclusion}

This work demonstrates that character interaction networks provide a powerful, language-independent representation of literary narratives. Learned graph representations significantly outperform traditional and unsupervised baselines, enabling effective author identification in Urdu novels and opening new directions for computational literary analysis.

\bibliographystyle{plain}
\bibliography{references}

\end{document}